# Deciphering and integrating invariants for neural operator learning with various physical mechanisms

Rui Zhang [1], Qi Meng [2,*] and Zhi-Ming Ma [1,*]

## ABSTRACT

Neural operators have been explored as surrogate models for simulating physical systems to overcome the limitations of traditional partial differential equation (PDE) solvers. However, most existing operator learning methods assume that the data originate from a single physical mechanism, limiting their applicability and performance in more realistic scenarios. To this end, we propose the physical invariant attention neural operator (PIANO) to decipher and integrate the physical invariants for operator learning from the PDE series with various physical mechanisms. PIANO employs self-supervised learning to extract physical knowledge and attention mechanisms to integrate them into dynamic convolutional layers. Compared to existing techniques, PIANO can reduce the relative error by 13.6%–82.2% on PDE forecasting tasks across varying coefficients, forces or boundary conditions. Additionally, varied downstream tasks reveal that the PI embeddings deciphered by PIANO align well with the underlying invariants in the PDE systems, verifying the physical significance of PIANO.

## INTRODUCTION

Partial differential equations (PDEs) provide a fundamental mathematical framework to describe a wide range of natural phenomena and physical processes, such as fluid dynamics [1], life science [2] and quantum mechanics [3], among others. Accurate and efficient solutions of PDEs are essential for understanding and predicting the behavior of these physical systems. However, due to the inherent complexity of PDEs, analytical solutions are often unattainable, necessitating the development of numerical methods for their approximation [4]. Over the years, numerous numerical techniques have been proposed for solving PDEs, such as the finite difference method, finite element method and spectral method [5]. These methods have been widely used in practice, providing valuable insights into the behavior of complex systems governed by PDEs [6,7]. Despite the success of classical numerical methods in solving a wide range of PDEs, there are several limitations associated with these techniques, such as the restriction on step size, difficulties in handling complex geometries and the curse of dimensionality for high-dimensional PDEs [8–10].

In recent years, machine learning (ML) methods have evolved as a disruptive technology to classical numerical methods for solving scientific calculation problems for PDEs. By leveraging the power of data-driven techniques or the expression ability of neural networks, ML-based methods have the potential to overcome some of the shortcomings of traditional numerical approaches [8,11–15]. In particular, by using the deep neural network to represent the solution of the PDE, ML methods can efficiently handle complex geometries and solve high-dimensional PDEs [16]. The representative works include the DeepBSDE method, which can solve parabolic PDEs in 100 dimensions [8]; the random feature model, which can easily handle complex geometries and achieve spectral accuracy [17]; the ML-based reduced-order modeling, which can improve the accuracy and efficiency of traditional reduced-order modeling for nonlinear problems [18–20]. However, these methods are applied to the fixed initial field (or external force field), and they require the retraining of neural networks when solving PDEs with changing high-dimensional initial fields.

[1] Academy of Mathematics and Systems Science, Chinese Academy of Sciences (CAS), Beijing 100190, China and [2] Microsoft Research, Beijing 100080, China

*Corresponding authors. E-mails: meq@microsoft.com, mazm@amt.ac.cn

In addition to these developments, neural operators have emerged as a more promising approach to simulate physical systems with deep learning, using neural networks as surrogate models to learn the PDE operator between functional spaces from data [9,21,22], which can significantly accelerate the simulation process. Most studies along this line focus on the network architecture design to ensure both simulation accuracy and inference efficiency. For example, DeepONet [21] and its variants [23–25], Fourier neural operators [9,26,27] and transformer-based operators [28,29] have been proposed to respectively deal with continuous input and output space different frequency components and complex geometries. Compared to traditional methods neural operators break the restriction on spatiotemporal discretization and enjoy a speed up of thousands of times, demonstrating enormous potential in areas such as inverse design and physical simulations, among others [9,30]. However, these methods only consider PDEs generated from a single formula by default, limiting the applicability of neural operators to multi-physical scenarios, e.g. datasets of the PDE systems sampled under different conditions (boundary conditions, parameters, etc.).

To address this issue, message-passing neural networks (MPNNs) incorporate the indicator of the scenario (i.e. the PDE parameters) into inputs to improve the generalization capabilities of the model [10]. DyAd supervision learns the physical information through an encoder and automatically adapts it to different scenarios [31]. Although incorporating the physical knowledge can enhance the performance of the neural operator these methods still require access to the high-level PDE information in the training or test stage [10,31]. However, in many real-world applications collecting high-level physical information that governs the behavior of PDE systems can be infeasible or prohibitively expensive. For example, in fluid dynamics or ocean engineering, scientists can gather numerous flow field data controlled by varying and unknown Reynolds numbers, and calculating them would require numerous calls to PDE solvers [12,32].

To this end, we propose the physical invariant attention neural operator (PIANO), a novel operator learning framework for deciphering and integrating physical knowledge from PDE series with various PIs, such as varying coefficients and boundary conditions. PIANO has two branches: a PI encoder that extracts physical invariants and a personalized operator that predicts the complementary field representation of each PDE system (Fig. 1(a)). As illustrated in Fig. 1, PIANO employs two key designs: the contrastive learning stage for learning the PI encoder and an attention mechanism to incorporate this knowledge into neural operators through dynamic convolutional (DyConv) layers [33]. On the one hand contrastive learning extracts the PI representation through the similarity loss defined on augmented spatiotemporal patches cropped from the dataset (Fig. 1(b)). To enhance consistency with physical priors we propose three physics-aware cropping techniques to adapt different PI properties for different PDE systems, such as spatiotemporal invariant, boundary invariant, etc. (Fig. 1(b)(iii)). This physics-aware contrastive learning technique extracts the PI representation without the need for the labels of the PDE conditions, thus providing the corresponding PI information for each PDE series (Fig. 1(b)). On the other hand, after the PI encoder is trained by contrastive learning, we compute attention (i.e. $a_k^i$ in Fig. 1(c)) of the PI representation extracted by the PI encoder and reweight the convolutional kernel in the DyConv layer to obtain a personalized operator (Fig. 1(c)). This personalized operator, incorporated with the PI information as an indicator of the PDE condition, can predict the evolution of each PDE field in a mixed dataset with guaranteed generalization performance.

We demonstrate our method's effectiveness and physical meaning on several benchmark problems, including Burgers' equation, the convection-diffusion equation (CDE) and the Navier-Stokes equation (NSE). Our results show that PIANO achieves superior accuracy and generalization compared to existing methods for solving PDEs with various physical mechanisms. According to the results of four experiments, PIANO can reduce the relative error rate by 13.6%–82.2% by deciphering and integrating the PIs of PDE systems. Furthermore, we conduct experiments to evaluate the quality of PI embedding through some downstream tasks, such as unsupervised dimensionality reduction and supervised classification (regression). These results indicate that the manifold structures of PI embeddings align well with the underlying PIs hidden in the PDE series (e.g. Reynolds numbers in NSE and external forces in Burgers' equation), thereby enjoying the physical significance.

## THE FRAMEWORK OF PIANO

In this section, we introduce the framework of PIANO, including how PIANO deciphers PIs from unlabeled multi-physical datasets and the procedure to incorporate them into the neural operator.

### Description of the PDE system

Consider the time-dependent PDE system, which can be expressed as

$$\begin{cases} \partial_t \mathrm{u} = \mathcal{R}(\mathrm{u}, \theta_\mathcal{R}), \\ \mathrm{u}(\mathbf{x}, 0) = \mathrm{u}_0(\mathbf{x}), \quad (\mathbf{x}, t) \in \Omega \times [0, T] \\ \mathcal{B}[\mathrm{u}](\mathbf{x}, t, \theta_\mathcal{B}) = 0, \end{cases}, \tag{1}$$

where $\mathcal{R}$ is the differential operator with parameter $\theta_\mathcal{R} \in \Theta_\mathcal{R}$, $\Omega$ is a bounded domain and $\mathrm{u}_0$ represents the initial conditions. Let $\mathcal{B}[\mathrm{u}]$ be the boundary condition governed by the parameter $\theta_\mathcal{B} \in \Theta_\mathcal{B}$. Let $\Theta := \Theta_\mathcal{R} \times \Theta_\mathcal{B}$ be the product space between $\Theta_\mathcal{R}$ and $\Theta_\mathcal{B}$, and let $\theta := (\theta_\mathcal{R}, \theta_\mathcal{B}) \in \Theta$ be the global parameters of the PDE system. We utilize $\mathbf{u}_{k,t}[\Omega] := [\mathrm{u}_k[\Omega], \ldots, \mathrm{u}_{k+t-1}[\Omega]]$ to denote the $t$ frame ($t \in \mathbb{N}^+$) PDE series defined in $\Omega$.

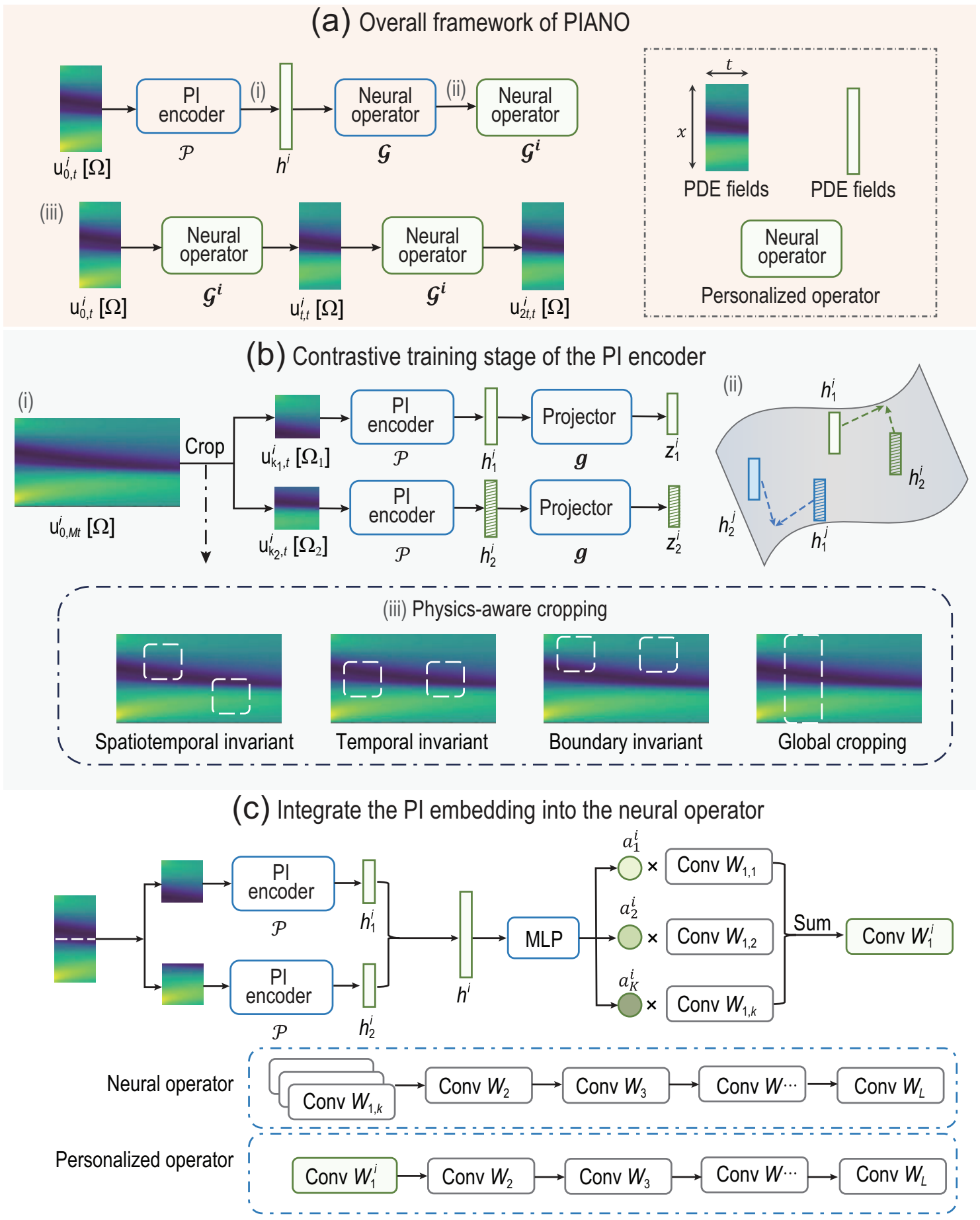

**Figure 1.** Illustration of PIANO. (a) The overall framework for PIANO when forecasting the PDE series. Given the $i$th PDE initial fields $\mathbf{u}^i_{0,t}[\Omega]$, PIANO first infers the PI embedding $\mathbf{h}^i$ via the PI encoder $\mathcal{P}$, and then integrates $\mathbf{h}^i$ into neural operator $\mathcal{G}$ to obtain a personalized operator $\mathcal{G}^i$ for $\mathbf{u}^i$. After that, PIANO predicts the subsequent PDE fields with this personalized operator. (b) Training stage of the PI encoder. (i) Illustration of contrastive learning. We crop two patches from each PDE series in a mini-batch according to the physical priors. The PI encoder and the projector are trained to maximize the similarity of two homologous patches. (ii) The effect of SimCLR loss, which brings closer (pushes apart) the representations governed by the same (different) physical parameters. (iii) Physics-aware cropping strategy of contrastive learning in PIANO. The cropping strategy should align with the physical prior of the PDE system. We illustrate the cropping strategies for spatiotemporal, temporal and boundary invariants. We also represent the global cropping strategy for comparison, which does not consider the more detailed physical priors and feeds the entire spatial fields directly. (c) Integration of the PI embedding into the neural operator. We use a split-merge trick to obtain the PI embedding $\mathbf{h}^i$ for the PDE field $\mathbf{u}^i$, and feed $\mathbf{h}^i$ into a multi-layer perception (MLP) to obtain $K$ non-negative scales $\{a^i_k\}^K_{k=1}$ with $\sum_k a^i_k = 1$. We use $a^i_k$ as the attention to reweight the DyConv layer in the neural operator and thus obtain a personalized operator for $\mathbf{u}^i$, which is incorporated with physical knowledge in $\mathbf{h}^i$.

In this paper, we consider the scenario where $\theta \in \Theta$ is a time-invariant parameter. In other words, the parameters $\theta$ that govern the PDE system in Equation (1) do not change over time, which includes the following three scenarios.

- Spatiotemporal invariant: $\mathbf{u}_{k_1,t}[\Omega_1]$ and $\mathbf{u}_{k_2,t}[\Omega_2]$ share the same $\theta_\mathcal{R}$ for all $k_1, k_2 \in [0, T]$ and $\Omega_1, \Omega_2 \subset \Omega$.
- Temporal invariant: given $\Omega' \subset \Omega$, $\mathbf{u}_{k_1,t}[\Omega']$ and $\mathbf{u}_{k_2,t}[\Omega']$ share the same $\theta_\mathcal{R}$ for all $k_1, k_2 \in [0, T]$.
- Boundary invariant: $\mathbf{u}_{k_1,t}[\Omega]$ and $\mathbf{u}_{k_2,t}[\Omega]$ share the same $\theta_\mathcal{B}$ for all $k_1, k_2 \in [0, T]$.

In Table 1, we give some examples of one-dimensional (1D) heat equations to illustrate the above three types of PI.

## The learning regime

Given the $t$ frame PDE series $\mathbf{u}_{k,t}[\Omega]$ governed by Equation (1), an auto-regressive neural operator $\mathcal{G}$ acts as a surrogate model, which produces the next $t$ frame PDE solution as follows:

$$\mathcal{G}(\mathbf{u}_{k,t}[\Omega]) = \mathbf{u}_{k+t,t}[\Omega]. \tag{2}$$

We assume that the neural operator $\mathcal{G}$ is trained under the supervision of the dataset $\mathbb{D}_{\text{train}} := \{\mathbf{u}^i_{0,Mt}[\Omega]\}^I_{i=1}$ ($M \in \mathbb{N}^+$), where $\mathbf{u}^i_{0,Mt}[\Omega]$ is the $i$th PDE series defined in $\Omega \times [0, Mt]$ and governed by the parameter $\theta^i \in \Theta$. Existing methods typically assume that all $\mathbf{u}^i$ in $\mathbb{D}_{\text{train}}$ share the same $\theta$ [9,21] or have known different parameters $\theta^i$ [10,31]. However, we consider a more challenging scenario where data are generated from various physical systems (with varying but unknown $\theta^i$ in $\mathbb{D}_{\text{train}}$ and $\mathbb{D}_{\text{test}}$) and no additional knowledge of $\theta^i$ is provided during the training and test stages.

## Forecasting stage of PIANO

As shown in Fig. 1(a), given the initial PDE fields $\mathbf{u}^i_{0,t}[\Omega]$, the forecasting stage of PIANO includes three steps: (1) *infer* the PI embedding $\mathbf{h}^i$ via the PI encoder $\mathcal{P}$; (2) *integrate* $\mathbf{h}^i$ into neural operator $\mathcal{G}$ to obtain a personalized operator $\mathcal{G}^i$ for $\mathbf{u}^i$; (3) *predict* the subsequent PDE fields with the personalized operator $\mathcal{G}^i$. As a result, two key technical problems arise when performing the above plans. On the one hand, we need to decipher the PI information behind the PDE system without the supervision of known labels. To this end, we utilize contrastive learning to pre-train the PI encoder in a self-supervised manner and propose the physics-aware cropping strategy to constrain the learned representation to align with the physical prior. On the other hand, we need to integrate the PI embedding into the neural operator to obtain the personalized operator. In this paper, we borrow the DyConv technique [33] and propose the split-merge trick to use the PI embedding fully.

**Table 1.** Examples of three types of PIs on 1D heat equations ($\Omega = [-1, 1]$).

| PDE formula | Type of PI | PI $\theta$ | PI space $\Theta$ |
|---|---|---|---|
| $\partial_t \mathrm{u} = \kappa \Delta \mathrm{u}, \mathrm{u}(\pm 1, t) = 0$ | Spatiotemporal invariant | $\kappa$ | $[0, 1]$ |
| $\partial_t \mathrm{u} = 0.1 \Delta \mathrm{u} + f(x), \mathrm{u}(\pm 1, t) = 0$ | Temporal invariant | $f(x)$ | $\{a \sin(x) : a \in [0, 1]\}$ |
| $\partial_t \mathrm{u} = 0.1 \Delta \mathrm{u}, \mathrm{u}(\pm 1, t) = c$ | Boundary invariant | $c$ | $[-1, 1]$ |

## Contrastive training stage of the PI encoder

In this section we introduce how to train an encoder $\mathcal{P}$ for extracting the PI information from the training set $\mathbb{D}_{\text{train}} = \{\mathbf{u}^i_{0,Mt}[\Omega]\}_{i=1}^{I}$ that is generated from various PDE fields without the supervision of $\{\theta^i\}_{i=1}^{I}$. We begin by considering the scenario where $\theta^i$ is a spatiotemporal invariant, i.e. $\mathbf{u}^i_{k_1,t}[\Omega_1]$ and $\mathbf{u}^i_{k_2,t}[\Omega_2]$ share the same $\theta^i$ for all $k_1, k_2 \in [0, T]$ and $\Omega_1, \Omega_2 \subset \Omega$. When $\theta \in \Theta$ is identifiable, there exists a mapping $\mathcal{M}$ satisfying the property

$$\mathcal{M}(\mathbf{u}^i_{k_1,t}[\Omega_1]) = \mathcal{M}(\mathbf{u}^i_{k_2,t}[\Omega_2]) = \theta^i. \tag{3}$$

However, the mapping $\mathcal{M}$ that can directly output $\theta^i$ is not available due to the absence of $\theta$. To decipher the information implying $\theta^i$, we adopt the technique from SimCLR [34] to train $\mathcal{P}$ in a self-supervised manner. In each mini-batch we sample training data $\{\mathbf{u}^i_{0,Mt}[\Omega]\}_{i\in\mathcal{A}}$ from $\mathbb{D}_{\text{train}}$ with index set $\mathcal{A}$ and randomly intercept two patches from each PDE sample, i.e. $\{\mathbf{u}^i_{k_1,t}[\Omega_1]\}_{i\in\mathcal{A}}$ and $\{\mathbf{u}^i_{k_2,t}[\Omega_2]\}_{i\in\mathcal{A}}$. The PI encoder $\mathcal{P}$ maps each patch to a representation vector, denoted $\mathbf{h}^v_i := \mathcal{P}(\mathbf{u}^i_{k_v,t}[\Omega_v])$ for $v \in \{1, 2\}$. Subsequently, we employ a two-layer MLP $g$ as a projection head to obtain $\mathbf{z}^v_i = g(\mathbf{h}^v_i)$ (Fig. 1(b)(i)). Considering the PDE patches cropped from the same/different PDE series as positive/negative samples, the SimCLR loss can be expressed as

$$\mathcal{L}_{\text{SimCLR}} = -\frac{1}{2|\mathcal{A}|} \sum_{i\in\mathcal{A}} \times \left[ \log \frac{\exp\left(\text{sim}\left(\mathbf{z}^i_1, \mathbf{z}^i_2\right)/\tau\right)}{\sum_{j\neq i}\sum_{v\in\{1,2\}} \exp\left(\text{sim}\left(\mathbf{z}^i_1, \mathbf{z}^j_v\right)/\tau\right)} + \log \frac{\exp\left(\text{sim}\left(\mathbf{z}^i_1, \mathbf{z}^i_2\right)/\tau\right)}{\sum_{j\neq i}\sum_{v\in\{1,2\}} \exp\left(\text{sim}\left(\mathbf{z}^i_2, \mathbf{z}^j_v\right)/\tau\right)} \right], \tag{4}$$

where $\text{sim}(\mathbf{u}, \mathbf{v}) := \mathbf{u}^\top \mathbf{v}/\|\mathbf{u}\|\|\mathbf{v}\|$ denotes the cosine similarity between $\mathbf{u}$ and $\mathbf{v}$, and $\tau > 0$ denotes a temperature parameter. As shown in Fig. 1(b)(ii), the SimCLR loss brings the representations governed by the same physical parameters closer, while pushing apart those with different parameters. After the training stage of contrastive learning, we throw away projector $g$ and only utilize encoder $\mathcal{P}$ to extract PI information from PDE fields, which is in line with the SimCLR method [34]. See the Method section for more details on the architecture of the PI encoder and the physics-aware cropping strategy.

## Integrate the PI representation

In this section we introduce how PIANO integrates the pre-trained PI representation into the neural operator. Given the pre-trained PI encoder $\mathcal{P}$ and an initial PDE field $\mathbf{u}^i_{0,t}[\Omega]$, we first obtain the PI embedding $\mathbf{h}^i$ via a split-merge trick (see the Method section for more details), and then we adopt the DyConv [33] technique to incorporate the PI information into the neural operator $\mathcal{G}$. In the first layer of $\mathcal{G}$ there are $K$ convolutional matrices of the same size, denoted $\{\mathbf{W}_{1,k}\}_{k=1}^{K}$. In detail, we transform the first Fourier or convolutional layer into a DyConv layer in the Fourier-based or convolutional-based neural operators, respectively. All other layers maintain the same structure as the original neural operators. When predicting the PDE fields for a specific instance $\mathbf{u}^i$, we use an MLP to transform its PI representation $\mathbf{h}^i$ into $K$ non-negative scales $\{a^i_k\}_{k=1}^{K}$ with $\sum_k a^i_k = 1$. The normalization of $a^i_k$ is implemented by a softmax layer. We use $\{a^i_k\}_{k=1}^{K}$ as the attention to reweight the $K$ convolution matrices, i.e. $\mathbf{W}^i_1 = \sum_k a^i_k \mathbf{W}_{1,k}$. We replace the first layer of $\mathcal{G}$ with $\mathbf{W}^i_1$ and denote this new operator as $\mathcal{G}^i$, which can be considered as the personalized operator for $\mathbf{u}^i$ (Fig. 1(c)). It is worth mentioning that the parameters $\mathbf{W}^i_1$ in $\mathcal{G}^i$ are obtained by the weighted summation, whose computational cost is almost negligible compared with the convolutional operation. Therefore, when aligning the parameters of PIANO and other neural operators, PIANO enjoys a comparable or faster inference speed, even considering the calculation of the PI representation $\mathbf{h}$.

## EXPERIMENTS

In this section, we conduct a series of numerical experiments to assess the performance of our proposed PIANO method and other baseline techniques in simulating PDE systems governed by diverse PIs.

### Experimental setup

We divide the temporal intervals into 200 frames for training and validation. The input and output frames are set as 20 for neural operator and PI encoders in the experiments. In order to assess the out-of-distribution generalization capabilities of the trained operator, we set the test temporal intervals at 240, with the last 40 frames occurring exclusively in the test set. We refer to the temporal interval in the training set as the training domain, and the temporal interval that only occurs in the test set as the future domain. The spatial intervals are partitioned into 64 frames for the 1D case and $64 \times 64$ frames for the 2D case. The training, test and validation set sizes for all tasks are 1000, 200 and 200, respectively. All experiments are carried out using the PyTorch package [35] on an NVIDIA A100 GPU. We repeat each experiment with three random seeds from the set {0 1, 2} and report the mean value and variance. The performance of the model is evaluated using the average relative $\ell_2$ error ($E_{\ell_2}$) and the $\ell_\infty$ error ($E_{\ell_\infty}$) over all frames in the training domain and the future domain, respectively.

### Dataset

In this section, we introduce the PDE dataset utilized in this paper, including two kinds of Burgers' equation, the 1D CDE and three kinds of 2D NSEs.

*Experiment E1: Burgers' equation with varying external forces f.* We simulate the 1D Burgers' equation with varying external forces $f$, defined as

$$\frac{\partial u}{\partial t} = -u\frac{\partial u}{\partial x} + 0.1\Delta u + 0.1 f(x),$$
$$x \in [-\pi, \pi], \quad u(\pm\pi, t) = 0, \tag{5}$$

where $f(x)$ is a smooth function representing the external force. In this experiment, we select 14 different $f$ to evaluate the performance of PIANO and other baseline methods under varying external forces. These forces are uniformly sampled from the set $\{0, 1, \cos(x), \cos(2x), \cos(3x), \sin(x), \sin(2x), \sin(3x), \pm\tanh(x), \pm\tanh(2x), \pm\tanh(3x)\}$. The ground-truth data are generated using the Python package 'py-pde' [36] with a fixed step size of $10^{-4}$. The final time $T$ is set to 5 for the training set and 6 for the test set.

*Experiment E2: Burgers' equation with varying diffusivities D.* We simulate the 1D Burgers' equation with spatially varying diffusivities, defined as

$$\frac{\partial u}{\partial t} = -u\frac{\partial u}{\partial x} + 0.1\nabla(D(x)\cdot\nabla u),$$
$$x \in [-\pi, \pi], \quad u(\pm\pi, t) = 0, \tag{6}$$

where $D(x)$ is a smooth and non-negative function representing the spatially varying diffusivity. In this experiment, we select 10 different diffusivities to evaluate the performance of PIANO and other baseline methods under varying spatial fields. Ten types of diffusivities are uniformly sampled from the set $\{1, 2, 1 \pm \cos(x), 1 \pm \sin(x), 1 \pm \cos(2x), 1 \pm \sin(2x)\}$. The data generation scheme and the final time $T$ are aligned with experiment E1.

*Experiment E3: CDE with varying boundary conditions $\mathcal{B}$.* We simulate the 1D CDEs with varying boundary conditions, defined as

$$\frac{\partial u}{\partial t} = 0.1\Delta u + 0.1u + 0.1\sin(2\pi x),$$
$$x \in [0, 1], \quad \mathcal{B}[u](x, t) = 0, \tag{7}$$

where $\mathcal{B}[u](x, t) = 0$ represents the boundary conditions. In this experiment, we select four types of $\mathcal{B}$ to evaluate the generalizability of PIANO and other baseline methods under varying boundary conditions. In this dataset, four types of boundary conditions include the Dirichlet condition ($u = 0.2$), the Neumann condition ($\partial_n u = 0.2$), the curvature condition ($\partial_n^2 u = 0.2$) and the Robin condition ($\partial_n u + u = 0.2$). The data generation scheme and the final time $T$ align with experiment E1.

*Experiment E4: NSE with varying viscosity terms $\nu$.* We simulate the vorticity fields for 2D flows within a periodic domain $\Omega = [0, 1] \times [0, 1]$, governed by the NSEs:

$$\frac{\partial \omega}{\partial t} = -(\mathbf{u}\cdot\nabla)\omega + \nu\Delta\omega + f(\mathbf{x}),$$
$$\omega = \nabla \times \mathbf{u}, \tag{8}$$

where $f(\mathbf{x}) = 0.1\sin(2\pi(\mathbf{x}_1 + \mathbf{x}_2)) + 0.1\cos(2\pi(\mathbf{x}_1 + \mathbf{x}_2))$ and $\nu \in \mathbb{R}^+$ represents the forcing function and viscosity term, respectively. The viscosity is a crucial component in NSEs that determines the turbulence of flows [37,38]. We generate NSE data with varying viscosity coefficients to simulate heterogeneity, ranging from $10^{-2}$ to $10^{-5}$. The viscosity fields become more complicated as $\nu$ decreases because the nonlinear term $-(\mathbf{u}\cdot\nabla)\omega$ gradually governs the motion of the fluids. The data generation process employs the pseudo-spectral method with a time step of $10^{-4}$ and a $256 \times 256$ grid size. The data are then downsampled to a grid size $64 \times 64$, which aligns with the settings in [9]. The final time $T$ is 20 and 24 for the training and test sets respectively.

*Experiment E5: NSE with varying viscosity terms $\nu$ and external forces $f$.* In this experiment, we aim to simulate the 2D NSE as shown in Equation (8), with varying viscosity terms $\nu$ and external forces $f$. The viscosity coefficients $\nu$ range from $10^{-2}$ to $10^{-5}$. The form of the forcing function is given by $f(\mathbf{x}) = a\sin(2\pi(\mathbf{x}_1 + \mathbf{x}_2)) + a\cos(2\pi(\mathbf{x}_1 + \mathbf{x}_2))$, where the coefficient $a$ is uniformly sampled from $[0, 0.2]$. All other experimental settings are consistent with those described in experiment E4.

*Experiment E6: Kolmogorov flow with varying viscosity terms $\nu$.* We simulate the vorticity fields for 2D NSEs within a periodic domain $\Omega = [0, 1] \times [0, 1]$ driven by Kolmogorov forcing [39]:

$$\frac{\partial \omega}{\partial t} = -(\mathbf{u} \cdot \nabla)\omega + \nu\Delta\omega + 0.1\cos(8\pi\mathbf{x}_1)$$
$$\omega = \nabla \times \mathbf{u}. \tag{9}$$

The fluid fields in Equation (9) result in much more complex trajectories due to the involvement of Kolmogorov forcing. We generate NSE data with varying viscosity coefficients to simulate heterogeneity, ranging from $10^{-2}$ to $10^{-4}$. All other experimental settings are consistent with those described in experiment E4.

## Baselines

We consider several representative baselines from operator learning models, including the following.

- Fourier neural operator (FNO) [9]: a classical neural operator that uses the Fourier transform to handle PDE information in the frequency domain.
- Unet [40,41]: a classic architecture for semantic segmentation in biomedical imaging recently utilized as a surrogate model for PDE solvers.
- Low-rank decomposition network (LordNet) [42]: a convolutional-based neural PDE solver that learns a low-rank decomposition layer to extract dominant patterns.
- MultiWaveleT- (MWT) based model [43]: a neural operator that compresses the kernel of the corresponding operator using a fine-grained wavelet transform.
- Factorized Fourier neural operators (FFNOs) [27]: an FNO variant that improves performance using a separable spectral layer and enhanced residual connections.

For PIANO we conduct experiments on PIANO + X, where X represents the backbone models. For the neural operator X and PIANO + X, we align the critical parameters of X and adjust the widths of the networks to match the number of parameters between X and PIANO + X, thereby ensuring a fair comparison.

## Results

Table 2 presents the performance of various models for the PDE simulation on the experiments (E1–E6), as well as their computational costs. PIANO achieves the best prediction results across most metrics and experiments. When compared with the backbone models X (FNO, Unet and FFNO), the three variants of PIANO + X consistently outperform their backbone models on all tasks for both $E_{\ell_2}$ and $E_{\ell_\infty}$ errors, demonstrating that the PI embedding can enhance the robustness and accuracy of neural operators' prediction capabilities. Specifically, PIANO + FNO, compared to FNO, reduces the error rate $E_{\ell_2}$ by 26.5%–63.1% in the training domain and by 35.7%–51.7% in the future domain over four experiments. PIANO + Unet, compared to Unet, reduces the error rate $E_{\ell_2}$ by 32.9%–76.8% in the training domain and by 36.7%–82.2% in the future domain over four experiments. PIANO provides a more significant enhancement to Unet than FNO in most tasks. One potential explanation is that the Fourier layer within the PI encoder introduces additional frequency domain information to the convolution-based Unet. In contrast, FNO is already based on a Fourier layer network. We compare the vorticity fields (in E4 and E6) predicted by FNO and PIANO + FNO from $T = 4$ to $T = 24$ in Fig. 2. Within the training domain, PIANO demonstrates a superior ability to capture the intricate details of fluid dynamics compared to FNO. As for the future domain, where supervised data are lacking, PIANO and FNO struggle to provide exact predictions in E4. However, PIANO still forecasts the corresponding trends of fluids more accurately than FNO.

Regarding computational costs, it is worth mentioning that the PI encoder is a significantly lighter network (0.053 and 0.184 million for the Burgers and NSE cases) compared to the neural operator. As a result, the inference time added by the PI encoder is generally negligible, which is 0.002 and 0.004 s for the Burgers and NSE data, respectively. Furthermore, in situations where the computational cost of the convolutional layers in the backbone is substantial, PIANO can considerably enhance the computation speed with the help of dynamic convolutional techniques. For example, PIANO can reduce the inference time by 24.2% and 32.3% for FNO and FFNO, respectively, when simulating 2D NSEs. More detailed discussions on computational costs are given in the online supplementary material.

**Table 2.** Results of the PDE simulation for experiments E1, E2, E3, E4, E5 and E6. Relative errors (%) and computational costs for baseline methods and PIANO. The computational cost and numbers of parameters for PIANO reported in this table consider both the expenses of the PI encoder and neural operator. The best results in each task are highlighted in bold.

| | | Training domain | | Future domain | | Time | | Param |
|---|---|---|---|---|---|---|---|---|
| Data | Model | $E_{\ell_2}$ (%) | $E_{\ell_\infty}$ (%) | $E_{\ell_2}$ (%) | $E_{\ell_\infty}$ (%) | Train (s) | Infer (s) | # (million) |
| E1 | FNO | 0.669 ± 0.124 | 0.978 ± 0.029 | 1.062 ± 0.039 | 1.340 ± 0.158 | 0.128 | 0.018 | 0.757 |
| Burgers' equation with varying | LordNet | 1.660 ± 0.058 | 2.406 ± 0.262 | 2.782 ± 0.111 | 3.529 ± 0.213 | 0.317 | 0.138 | 0.810 |
| external forces $f$ | MWT | 1.962 ± 0.250 | 2.737 ± 0.450 | 2.764 ± 0.379 | 3.572 ± 0.514 | 0.460 | 0.111 | 0.789 |
| | Unet | 2.576 ± 0.124 | 4.205 ± 0.108 | 3.280 ± 0.084 | 4.687 ± 0.158 | 0.256 | 0.041 | 0.860 |
| | PIANO + FNO | **0.492 ± 0.045** | **0.611 ± 0.045** | **0.536 ± 0.046** | **0.700 ± 0.038** | 0.147 | 0.022 | 0.762 |
| | PIANO + Unet | 1.605 ± 0.264 | 3.130 ± 0.685 | 1.796 ± 0.386 | 2.946 ± 0.526 | 0.299 | 0.039 | 0.766 |
| E2 | FNO | 6.328 ± 0.162 | 10.847 ± 0.251 | 13.111 ± 0.384 | 19.379 ± 0.649 | 0.128 | 0.018 | 0.757 |
| Burgers' equation with varying | LordNet | 8.471 ± 0.628 | 22.016 ± 6.849 | 23.786 ± 7.989 | 62.977 ± 35.304 | 0.317 | 0.138 | 0.810 |
| diffusivities $D$ | MWT | 6.381 ± 0.069 | 12.355 ± 0.580 | 12.013 ± 0.266 | 18.952 ± 1.082 | 0.460 | 0.111 | 0.789 |
| | Unet | 7.087 ± 1.680 | 12.592 ± 2.750 | 13.593 ± 3.413 | 20.221 ± 5.280 | 0.256 | 0.041 | 0.860 |
| | PIANO + FNO | 4.559 ± 0.092 | 8.932 ± 0.312 | 8.421 ± 0.440 | 13.680 ± 1.174 | 0.147 | 0.022 | 0.762 |
| | PIANO + Unet | **4.149 ± 0.985** | **8.879 ± 1.106** | **7.342 ± 2.072** | **12.330 ± 3.015** | 0.299 | 0.039 | 0.766 |
| E3 | FNO | 1.127 ± 0.256 | 1.742 ± 0.346 | 1.468 ± 0.394 | 2.041 ± 0.420 | 0.128 | 0.018 | 0.757 |
| CDE with varying boundary | LordNet | 0.605 ± 0.039 | 0.990 ± 0.048 | 0.901 ± 0.072 | **0.832 ± 0.063** | 0.317 | 0.138 | 0.810 |
| conditions $\mathcal{B}$ | MWT | 0.662 ± 0.037 | 1.232 ± 0.107 | 0.781 ± 0.113 | 1.385 ± 0.148 | 0.460 | 0.111 | 0.789 |
| | Unet | 12.565 ± 1.752 | 20.786 ± 2.976 | 20.335 ± 3.100 | 22.686 ± 3.511 | 0.256 | 0.041 | 0.860 |
| | PIANO + FNO | **0.416 ± 0.180** | **0.893 ± 0.338** | **0.708 ± 0.403** | 1.098 ± 0.547 | 0.148 | 0.022 | 0.763 |
| | PIANO + Unet | 2.921 ± 0.363 | 5.773 ± 0.767 | 3.611 ± 0.830 | 5.446 ± 0.676 | 0.299 | 0.039 | 0.767 |
| E4 | FNO | 10.433 ± 0.298 | 16.937 ± 0.302 | 30.702 ± 1.043 | 56.563 ± 0.949 | 0.384 | 0.182 | 2.085 |
| NSE with varying viscosity | LordNet | 8.469 ± 0.633 | 15.574 ± 0.863 | 30.348 ± 0.838 | 57.728 ± 1.514 | 1.031 | 0.547 | 2.069 |
| terms $\nu$ | MWT | 10.135 ± 0.346 | 17.917 ± 0.253 | 32.232 ± 0.713 | 61.572 ± 1.487 | 1.067 | 0.229 | 2.295 |
| | Unet | 9.054 ± 0.204 | 18.483 ± 0.381 | 31.830 ± 0.496 | 60.106 ± 0.299 | 0.335 | 0.089 | 3.038 |
| | FFNO | 3.698 ± 0.160 | 6.943 ± 0.214 | 15.845 ± 0.572 | 35.766 ± 1.069 | 1.964 | 1.008 | 2.013 |
| | PIANO + FNO | 4.652 ± 0.396 | 9.191 ± 0.605 | 17.393 ± 0.672 | 39.953 ± 1.107 | 0.395 | 0.138 | 2.020 |
| | PIANO + Unet | 6.070 ± 0.397 | 15.356 ± 0.914 | 20.132 ± 1.288 | 47.079 ± 2.144 | 0.440 | 0.111 | 1.941 |
| | PIANO + FFNO | **3.140 ± 0.100** | **5.935 ± 0.098** | **12.155 ± 0.237** | **28.985 ± 0.456** | 1.364 | 0.682 | 1.888 |
| E5 | FNO | 19.277 ± 0.762 | 26.354 ± 0.848 | 44.467 ± 2.005 | 57.912 ± 1.934 | 0.384 | 0.182 | 2.085 |
| NSE with varying viscosity | LordNet | 27.675 ± 4.095 | 39.617 ± 7.149 | 76.273 ± 19.280 | 111.628 ± 17.546 | 1.031 | 0.547 | 2.069 |
| terms $\nu$ and external forces $f$ | MWT | 18.908 ± 0.768 | 25.361 ± 0.764 | 40.919 ± 1.317 | 53.123 ± 1.087 | 1.067 | 0.229 | 2.295 |
| | Unet | 25.374 ± 0.321 | 37.916 ± 0.260 | 52.505 ± 4.859 | 73.183 ± 6.822 | 0.335 | 0.089 | 3.038 |
| | FFNO | 8.032 ± 0.575 | 11.607 ± 0.781 | 20.750 ± 1.188 | 28.939 ± 1.652 | 1.964 | 1.008 | 2.013 |
| | PIANO + FNO | 9.082 ± 0.238 | 12.731 ± 0.525 | 21.795 ± 0.833 | 29.912 ± 1.176 | 0.457 | 0.144 | 2.071 |
| | PIANO + Unet | 12.829 ± 0.440 | 23.184 ± 1.812 | 24.060 ± 1.081 | 40.415 ± 1.803 | 0.491 | 0.115 | 2.158 |
| | PIANO + FFNO | **6.937 ± 0.199** | **9.736 ± 0.215** | **18.062 ± 0.913** | **25.411 ± 0.920** | 1.424 | 0.686 | 1.997 |
| E6 | FNO | 4.017 ± 0.101 | 5.250 ± 0.171 | 5.241 ± 0.027 | 6.842 ± 0.219 | 0.384 | 0.182 | 2.085 |
| Kolmogorov flow with varying | LordNet | 6.559 ± 0.969 | 8.159 ± 2.259 | 11.343 ± 1.448 | 17.940 ± 8.683 | 1.031 | 0.547 | 2.069 |
| viscosity terms $\nu$ | MWT | 4.663 ± 0.285 | 5.769 ± 0.350 | 6.511 ± 0.103 | 8.062 ± 0.272 | 1.067 | 0.229 | 2.295 |
| | Unet | 9.807 ± 2.673 | 19.449 ± 6.144 | 13.949 ± 3.593 | 27.505 ± 9.510 | 0.335 | 0.089 | 3.038 |
| | FFNO | 1.727 ± 0.050 | 2.194 ± 0.052 | 2.608 ± 0.067 | 3.357 ± 0.050 | 1.964 | 1.008 | 2.013 |
| | PIANO + FNO | 1.908 ± 0.074 | 2.419 ± 0.040 | 2.840 ± 0.126 | 3.552 ± 0.126 | 0.395 | 0.138 | 2.020 |
| | PIANO + Unet | 6.704 ± 0.201 | 12.143 ± 0.119 | 9.676 ± 0.248 | 16.495 ± 0.168 | 0.440 | 0.111 | 1.941 |
| | PIANO + FFNO | **1.491 ± 0.037** | **1.876 ± 0.023** | **2.277 ± 0.110** | **3.040 ± 0.155** | 1.364 | 0.682 | 1.888 |

## Physical explanation of the PI encoder

In this section, we describe experiments to investigate the physical significance of the PI encoder on the Burgers (E1) and NSE (E4) data; specifically, whether the learned representation can reflect the PI information hidden within the PDE system. We consider two kinds of downstream task, unsupervised dimensionality reduction and supervised classification (regression), to analyze the properties of PI embeddings for PIANO. Furthermore, we compare several corresponding baselines to study the effects of each component in PIANO as follows.

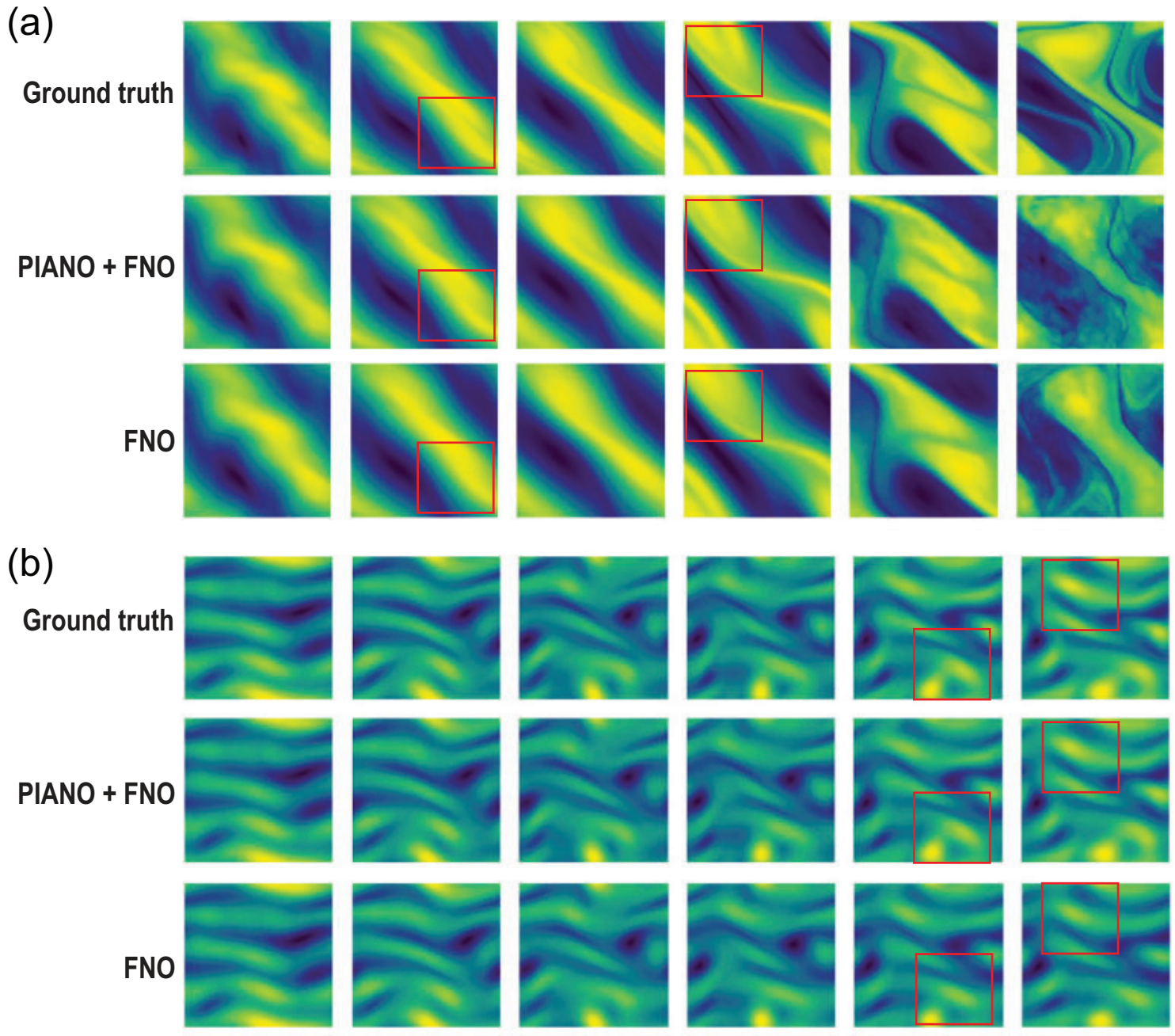

**Figure 2.** Comparison of the vorticity fields in E4 (a) and E6 (b) between FNO and PIANO + FNO from $T = 4$ to 24 in the periodic domain $[0, 1]^2$ for a 2D turbulent flow. Note that the times $T = \{4, 8, 12, 16, 20\}$ are in the training domain, while $T = 24$ is in the future domain. The vorticity fields in the bounding boxes indicate that PIANO can capture more details than FNO.

- PIANO-$\cancel{\mathcal{CL}}$: in this model, we jointly train the PI encoder and neural operator without the contrastive pre-training, which can be regarded as an FNO version for DyConv technique. We train this model to reveal the impact of contrastive learning in PIANO.
- PIANO-$\cancel{\mathcal{SM}}$: in PIANO, we utilize the split-merge trick to divide the PDE fields $\Omega$ into several patches $\{\Omega_v\}_{v=1}^{V}$ and then input them into the PI encoder during the training and testing phases (Fig. 1(b) and (c)). In PIANO-$\cancel{\mathcal{SM}}$, we directly feed the entire PDE fields into the PI encoder.
- PIANO-$\cancel{\mathcal{PC}}$: we assert that cropping strategies should align with the physics prior of the PDE system and propose physics-aware cropping methods for contrastive learning (Fig. 1(b)). In PIANO-$\cancel{\mathcal{PC}}$, we discard the physics-aware cropping technique and swap two corresponding augmentation methods for the Burgers and NSE data, respectively.

For dimensionality reduction tasks, we utilize UMAP [44] to project the PI embedding into a 2D and 1D manifold for the Burgers and NSE data respectively (Fig. 3). For Burgers' data, PIANO-$\cancel{\mathcal{CL}}$ fails to obtain a meaningful representation, highlighting the importance of contrastive learning. PIANO-$\cancel{\mathcal{SM}}$ and PIANO-$\cancel{\mathcal{PC}}$ can distinguish half of the external force types, but struggle to separate some similar functions, such as $-\tanh(kx)$ for $k \in \{1, 2, 3\}$. Only PIANO achieves remarkable clustering results (Fig. 3(a)). We also calculate four clustering metrics to quantitatively evaluate the performance of clustering (Fig. 3(b)), where the clustering results are obtained via K-means [45] with the PI representation. These four metrics include the silhouette coefficient, the adjusted Rand index, normalized mutual information and the Fowlkes–Mallows index, which assess the clustering quality through measuring intra-cluster similarity, agreement between partitions, shared information between partitions and the similarity of pairs within clusters, respectively. The larger their values, the better the clustering quality. As shown in Fig. 3(b), PIANO is the only method that achieves a silhouette coefficient greater than 0.65, with the other three metrics achieving values larger than 0.90; thus, PIANO significantly outperforms the other methods. For NSE data, PIANO is the only method where the first component of PI embeddings exhibits a strong correlation with the logarithmic viscosity term (with correlation coefficients greater than 98%). At the same time, the other three PIANO variants fail to distinguish viscosity terms ranging from $10^{-3}$ to $10^{-5}$ (Fig. 3(b)).

For supervised tasks, we train a linear predictor $\mathcal{T}$ that maps the learned representation $\mathbf{h}^i$ to the corresponding PDE parameters $\theta^i$ under the supervision of ground-truth labels (Table 3). For the dataset of Burgers' equation, which involves 14 types of external forces, the training of $\mathcal{T}$ naturally becomes a softmax regression problem. In the case of NSE, where the viscosity term continuously changes, we treat the training of $\mathcal{T}$ as a ridge regression problem. According to the supervised downstream tasks, the PI encoder trained in PIANO exhibits the best ability to predict the PIs in Burgers' equation and NSE compared to other baseline methods, which aligns with the experimental result in the unsupervised part.

The results of downstream tasks indicate that PIANO can represent the physical knowledge via a low-dimensional manifold and predict corresponding PDE parameters, thus demonstrating the physical meaning of PIANO.

## CONCLUSION

In this paper, we introduce PIANO, an innovative operator learning framework designed to decipher the PI information from PDE series with various physical mechanisms and integrate them into neural operators to conduct forecasting tasks. We propose the physics-aware cropping technique to enhance consistency with physical priors and the split-merge trick to fully utilize the physical information across the spatial domain. According to our numerical results, PIANO successfully overcomes the limitations of current neural operator learning methods, thereby demonstrating its capability to process PDE data from a diverse range of sources and scenarios. Furthermore, the results of a series of downstream tasks verify the physical significance of the extracted PI representation by PIANO.

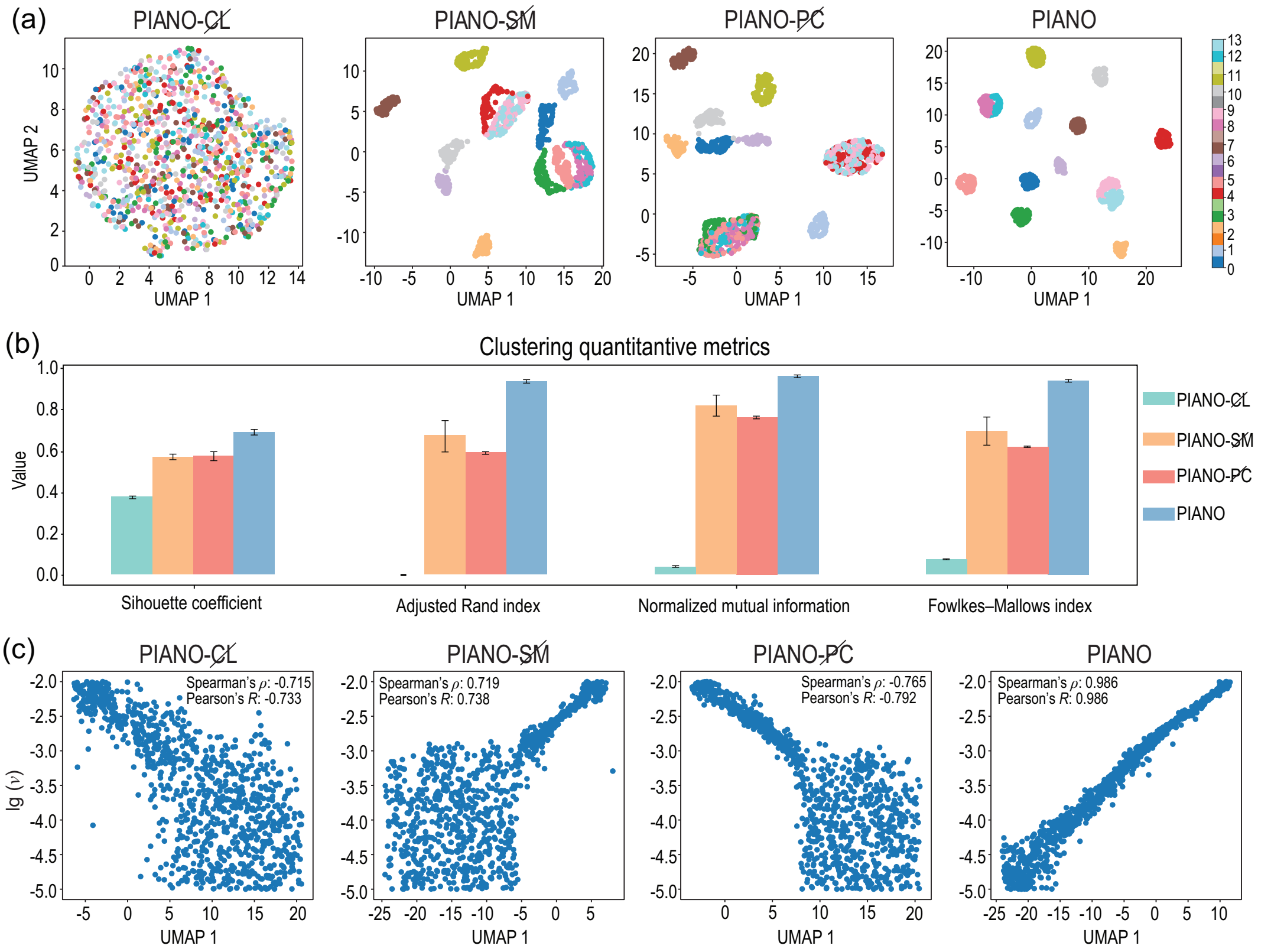

**Figure 3.** The performances of the learned representation on the unsupervised dimensionality reduction tasks. CL, SM and PC denote contrastive learning, the split-merge trick and the physics-aware cropping strategy, respectively. (a) The dimensionality reduction results of PI embeddings via UMAP for Burgers' data. The horizontal and vertical axes represent the two main components of UMAP, and each color represents a different external force in the dataset. Colors numbered from 0 to 13 correspond to the 14 types of external forces, including 0, 1, cos ($x$), sin ($x$), $-$tanh ($x$), tanh ($x$), cos (2$x$), sin (2$x$), tanh (2$x$), $-$tanh (2$x$), cos (3$x$), sin (3$x$), tanh (3$x$) and $-$tanh (3$x$). (b) Four metrics to evaluate the quantity of clustering via representation vectors given by different methods, including the silhouette coefficient, the adjusted Rand index, normalized mutual information and the Fowlkes–Mallows index. All four of these metrics indicate that the larger the value, the better the clustering performance. (c) The dimensionality reduction results of the PI embeddings via UMAP for the NSE data. The horizontal axis and the vertical axis represent the first component of UMAP and the logarithmic viscosity term lg($\nu$) in the dataset. We also calculate the Spearman and Pearson correlation coefficients between the first component and logarithmic viscosity term lg($\nu$), which represent the rank order and linear relationships between two variables, respectively.

**Table 3.** The performances of the learned representation on the supervised tasks. Accuracy (relative $\ell_2$ error) of the PI encoder in PIANO and other baselines using linear evaluation on Burgers' equation (NSE). CL, SM and PC denote contrastive learning, the split-merge trick and the physics-aware cropping strategy, respectively. The best results in each task are highlighted in bold.

| Method | Burgers' equation (accuracy, ↑) | NSE ($\ell_2$ error, ↓) |
|---|---|---|
| PIANO-~~CL~~ | 0.078 ± 0.003 | 0.161 ± 0.034 |
| PIANO-~~SM~~ | 0.988 ± 0.008 | 0.086 ± 0.011 |
| PIANO-~~PC~~ | 0.955 ± 0.013 | 0.092 ± 0.004 |
| PIANO | **0.997 ± 0.003** | **0.033 ± 0.002** |

We propose the following future works to enhance the capabilities and applications of PIANO further.

- *Expanding PIANO to PDE types with varying geometries.* In this study, we primarily focused on 1D equations when simulating PDEs with varying boundary conditions. However, it would be valuable to explore the extension of PIANO to more complex PDEs, such as PDEs with 2D and 3D complex geometries.

- *Addressing large-scale challenges using PIANO.* In large-scale real-world problems, such as weather forecasting, PIANO can potentially extract meaningful PI representations, such as geographical information of various regions. This capability could enhance the accuracy and reliability of forecasting tasks and other large-scale applications.
- *Integrating additional physical priors into PIANO.* Our current study assumes that the underlying PI in the PDE system is time invariant. However, real-world systems often exhibit other physical properties, such as periodicity and spatial invariance. By incorporating these additional physical priors into the contrastive learning stage, PIANO could be applied to a broader range of problems.

# METHOD

## Architecture of the PI encoder

In this paper, the architecture of $\mathcal{P}$ consists of six layers, successively including two Fourier layers [9] two convolutional layers and two fully connected layers. The Fourier layers can extract the PDE information in the frequency space and other layers downsample the feature map to a low-dimensional vector. We employ the 'GeLU' function as the activation function. It is important to note that we only feed a sub-patch of the PDE field to $\mathcal{P}$ and that the output of $\mathcal{P}$ is a low-dimensional vector. Furthermore the amount of information required to infer PIs is significantly less than that needed to forecast the physical fields in a PDE system. Consequently, compared with the main branch of the neural operator, this component is a lightweight network that extracts PIs and enjoys fast inference speed.

## Physics-aware cropping strategy

The cropping of the PDE series can be interpreted as data augmentation in contrastive learning. Unlike previous argumentation methods in vision tasks [46–49], those for PDE representation should comply with the physical prior accordingly. We have previously discussed cases where the PI represents spatiotemporal invariants. When the PI is only a temporal invariant and exhibits spatial variation, such as an external force, it is necessary to align spatial positions when implementing the crop operator. As a result, we extract two patches from the same spatial location for each PDE sample, i.e. $\{\mathbf{u}^i_{k_1,t}[\Omega^i]\}_{i\in\mathcal{A}}$ and $\{\mathbf{u}^i_{k_2,t}[\Omega^i]\}_{i\in\mathcal{A}}$. For boundary invariants, we need to crop the PDE patches near the boundary to encode the boundary conditions. We illustrate all three cropping methods in Fig. 1(b)(iii). Note that we also illustrate another cropping approach, called the global cropping technique, which directly selects the PDE patch across the entire spatial field as augmentation samples, i.e. $\{\mathbf{u}^i_{k_1,t}[\Omega]\}_{i\in\mathcal{A}}$ and $\{\mathbf{u}^i_{k_2,t}[\Omega]\}_{i\in\mathcal{A}}$. This global cropping strategy considers the time-invariant property of PIs, while ignoring the more detailed physical priors of different types of PI.

## Split-merge trick

We split the PDE fields according to the physical prior in the contrastive training stage. Compared to global cropping, such a splitting strategy can encode the physical knowledge into $\mathcal{P}$ through a more accurate approach. In the forecasting stage, we split the initial PDE fields $\mathbf{u}^i_{0,t}[\Omega]$ into $V$ uniform and disjointed patches $\{\mathbf{u}^i_{0,t}[\Omega_v]\}_{v=1}^V$, which are aligned with the patch size in the pre-training stage and satisfy $\cup_v \Omega_v = \Omega$. We feed all patches into $\mathcal{P}$ to obtain the corresponding representations $\mathbf{h}^i_v = \mathcal{P}(\mathbf{u}^i_{0,t}[\Omega_v])$, and merge them together as the PI vector of $\mathbf{u}^i$, i.e. $\mathbf{h}^i := [\mathbf{h}^i_1, \ldots, \mathbf{h}^i_V]$ (Fig. 1(c)). This merge operation can make full use of the PDE information. In practice, we fix the parameters of the pre-trained PI encoder $\mathcal{P}$ and only optimize the neural operator $\mathcal{G}$ in the training stage.

# MATERIALS AND ADDITIONAL EXPERIMENTS

Detailed experimental settings and additional experiments are reported in the online supplementary material. The source code is publicly available at https://github.com/optray/PIANO.

# SUPPLEMENTARY DATA

Supplementary data are available at *NSR* online.

# FUNDING

This work was supported by the National Key R&D Program of China (2020YFA0712700).

# AUTHOR CONTRIBUTIONS

R.Z. contributed to the primary idea, software designs, experiments and manuscript writing. Q.M. contributed to the original idea, supervised the whole project and revised the paper. Q.M. and Z.-M.M. led the related project and directed the study.

***Conflict of interest statement***. None declared.